\documentclass{article}
\usepackage{spconf,amsmath,graphicx}
\usepackage{booktabs}
\usepackage{adjustbox}
\usepackage{comment}
\usepackage{amsmath}
\usepackage{pifont}
\usepackage{xcolor}
\usepackage[normalem]{ulem}
\usepackage{caption}

\title{A$^{2}$V-SLP: Alignment-Aware Variational Modeling for Disentangled Sign Language Production}

\name{Sümeyye Meryem Taşyürek \qquad Enis Mücahid İskender \qquad Hacer Yalim Keles}
\address{Hacettepe University\\
Computer Engineering Department\\
Ankara, Türkiye}

\begin{document}
%
\maketitle
\begin{abstract}

Building upon recent structural disentanglement frameworks for sign language production, we propose A$^{2}$V-SLP, an alignment-aware variational framework that learns articulator-wise disentangled latent distributions rather than deterministic embeddings. A disentangled Variational Autoencoder (VAE) encodes ground-truth sign pose sequences and extracts articulator-specific mean and variance vectors, which are used as distributional supervision for training a non-autoregressive Transformer. Given text embeddings, the Transformer predicts both latent means and log-variances, while the VAE decoder reconstructs the final sign pose sequences through stochastic sampling at the decoding stage. This formulation maintains articulator-level representations by avoiding deterministic latent collapse through distributional latent modeling. In addition, we integrate a gloss attention mechanism to strengthen alignment between linguistic input and articulated motion. Experimental results show consistent gains over deterministic latent regression, achieving state-of-the-art back-translation performance and improved motion realism in a fully gloss-free setting.


\end{abstract}
\section{Introduction}
\label{sec:intro}

Sign Language Production (SLP) aims to generate continuous sign pose sequences from spoken language text, enabling accessible communication for Deaf and Hard-of-Hearing (DHH) communities. Unlike isolated sign synthesis, continuous SLP requires modeling complex spatio-temporal dynamics across multiple articulators. Despite recent progress, producing expressive and realistic sign sequences remains challenging.



Although gloss annotations are widely used in SLP, they are costly, difficult to standardize, and limited, motivating growing interest in gloss-free sign language production that generates pose sequences directly from text. However, temporal alignment between language and articulated motion remains a key challenge in gloss-free settings. Yin et al.~\cite{yin2023gloss} introduce gloss attention for sign language translation, which constrains attention to local neighborhoods and partially replaces the alignment role of gloss without explicit labels, making it directly relevant to SLP.


Recent advances in SLP adopt pose-based representations, where sign language is modeled as a sequence of skeletal keypoints rather than full video frames. This abstraction offers reduced computational cost, and greater flexibility for downstream rendering \cite{RASTGOO2024122846}. However, many approaches produce over smoothed motion, lose fine-grained hand articulation, or exhibit unnatural transitions between signs that significantly affect intelligibility and perceived naturalness. To address these issues adversarial methods, mixture density networks (MDN), vector quantization methods has been proposed but the problem of regression to the mean poses still persists \cite{walsh2024signstitchingnovelapproach, ma-etal-2024-multi}.

Recent work \cite{tasyurek2025disentangle, Kiziltepe_2025_ICCV} explores articulator-aware latent representations that separate hands, face, and body, improving manual articulation over holistic pose regression. While latent spaces offer compact and stable formulations, existing approaches rely on deterministic targets optimized with point-wise losses, which may bias learning toward central tendencies when multiple valid realizations exist. How articulatory variability is represented within such latent spaces remains an open question.

Motivated by these observations, we revisit gloss-free sign language production from representation and alignment perspectives. Building on articulator-aware latent modeling, we employ a structurally disentangled variational formulation that predicts latent mean and variance instead of deterministic embeddings. In addition, we integrate gloss attention to guide temporal alignment without relying on explicit gloss annotations. Together, this design combines structural disentanglement and alignment-aware attention to better capture articulatory variability while maintaining efficient and stable pose generation. Main contributions of this work are:


\begin{itemize}
    \item We introduce a \textbf{distributional supervision} for gloss-free sign language production, where articulator-wise latent mean and variance parameters extracted from a trained disentangled VAE are used as supervision targets. This replaces deterministic latent embeddings and mitigates latent collapse toward central values.
    \item We integrate \textbf{gloss attention} into continuous sign language production, demonstrating that alignment-aware attention can effectively support temporal correspondence between text input and articulated motion without relying on explicit gloss annotations. To the best of our knowledge, this is the first work to incorporate gloss attention into SLP models.
\end{itemize}

\section{RELATED WORK}
\label{sec:prior}

\subsection{SLP with Pose-Based Representations} 

Sign language production aims to generate sign sequences from spoken language text. Recent studies increasingly adopt pose-based representations, where sign language is modeled as sequences of skeletal keypoints while photorealistic video synthesis from these poses remains a separate challenge \cite{RASTGOO2024122846}. Pose-based SLP enables scalable modeling of continuous sign articulation. 

A central challenge in SLP is regression to the mean, where models generate averaged, over-smoothed motions, especially degrading hand and finger articulation, leading to reduced expressiveness and semantic drift, hindering communication quality \cite{walsh2024signstitchingnovelapproach, ma-etal-2024-multi}. Early work by Stoll et al. \cite{StollStephanie20180903} used neural machine translation to map text to glosses, but failed to produce continuous, natural sign sequences . Saunders et al. \cite{DBLP:journals/corr/abs-2004-14874} introduced a transformer-based SLP model, but its autoregressive decoding amplified the regression-to-the-mean effect. To mitigate this, Saunders et al. \cite{DBLP:journals/corr/abs-2008-12405} incorporated adversarial training with a multichannel sign production framework. Although this approach systematically addressed the regression-to-the-mean problem for the first time, it did not fully resolve it. Recently, Sign-IDD \cite{tang2024signiddiconicitydisentangleddiffusion} employs a 4D bone representation with diffusion-based generation.

Later works explored mixture density networks (MDN), combined with data augmentation and adversarial training techniques \cite{DBLP:journals/corr/abs-2103-06982}. Despite these efforts, user evaluations indicate that the regression-to-the-mean problem persists.  Ma et al.  \cite{ma-etal-2024-multi} tackled the regression-to-the-mean problem by introducing a dual-decoder transformer, separating manual and full-body decoding. Although this improved hand modeling, the approach suffered from high computational cost and sensitivity to data quality, limiting its scalability and generalization.


\subsection{Latent Space Modeling for SLP} 


Hwang et al. introduced NSLP-G, which generates sign poses in a Gaussian latent space using a VAE \cite{hwang2021non}. While this formulation addressed limitations of autoregressive decoding, it struggled with sequence length prediction and fine-grained hand articulation. Subsequent extensions improved low-variance detail modeling through length regulation and alternative reconstruction losses, but manual articulation remained inadequately represented \cite{hwang2022nonautoregressive}.


He et al. propose a latent pose dynamics framework for seamless sign language generation that decouples motion semantics from signer identity \cite{he2025motionchoreographerlearninglatent}. Recent work has explored structured and disentangled latent representations that explicitly separate articulators such as hands, face, and body, enabling selective weighting and region-specific modeling \cite{tasyurek2025disentangle}. This formulation introduces articulator-aware inductive biases and improves motion realism compared to direct pose regression. Iterative refinement strategies have been explored to progressively update latent representations during generation, improving robustness and temporal consistency without modifying the underlying latent abstraction \cite{Kiziltepe_2025_ICCV}. However, these approaches rely on deterministic latent supervision, leaving articulatory variability unmodeled. 


\subsection{Gloss-Free SLP and Alignment} 


Although gloss-based SLP models benefit from explicit intermediate supervision, gloss annotations are costly, inconsistent, and difficult to scale. This has motivated growing interest in gloss-free sign language production, which generates sign sequences directly from spoken language text \cite{hwang2021non, hwang2022nonautoregressive, hwang2024autoregressivesignlanguageproduction, walsh2024datadrivenrepresentationsignlanguage, tang2024signiddiconicitydisentangleddiffusion, tasyurek2025disentangle, Kiziltepe_2025_ICCV}. Discrete tokenization approaches based on vector quantization have been proposed to reduce gloss dependency, but often suffer from unnatural transitions and loss of fine-grained motion detail \cite{walsh2024signstitchingnovelapproach}.



A key difficulty in gloss-free settings is the absence of explicit temporal alignment between linguistic units and continuous sign motion. Yin et al.~\cite{yin2023gloss} show that gloss supervision implicitly enforces local semantic alignment, where each gloss mainly corresponds to a short span of neighboring frames. To approximate this behavior without gloss labels, they introduce gloss attention, a modified self-attention mechanism that restricts each query to attend only to a local temporal window of size $N$ rather than the full sequence. The attention window is centered around the current timestep and can include small learned offsets, allowing limited flexibility while preserving temporal locality. Although introduced for translation, gloss attention offers a relevant alignment inductive bias for sign language production.

\section{METHODOLOGY}
\label{sec:method}

\subsection{Structurally Disentangled Variational Latent Representation}

\begin{figure*}[h]
  \centering
  \includegraphics[width=16cm]{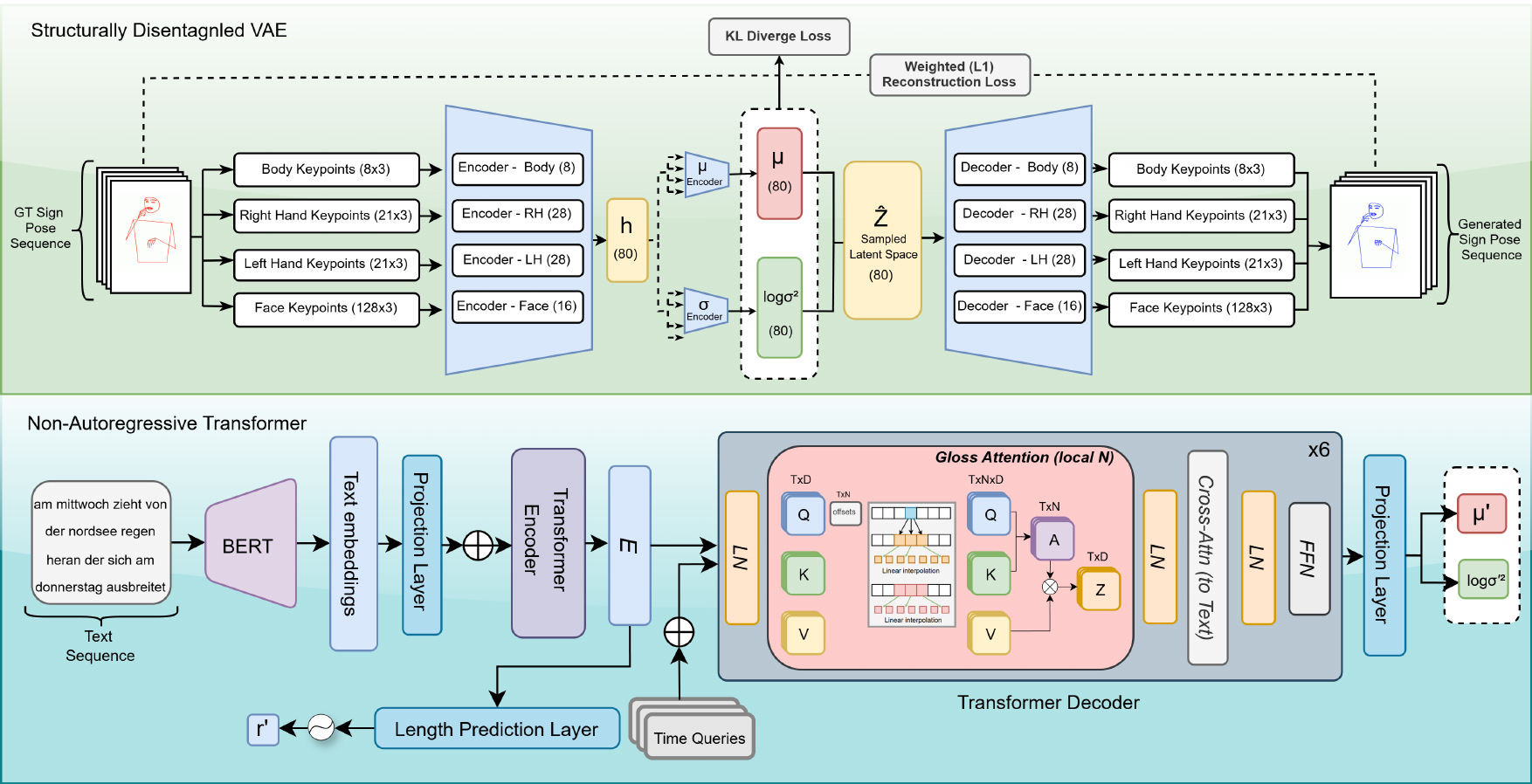}
  \caption{Overview of the proposed A$^{2}$V-SLP framework.
  BERT-based text embeddings are mapped by a non-autoregressive Transformer to structurally disentangled latent mean and variance parameters. During training, the pretrained VAE encoder provides articulator-wise latent distributions from ground-truth poses as supervision targets. Gloss attention replaces decoder self-attention with local temporal modeling, while cross-attention to text remains global. At inference time, latent samples are drawn from the predicted distributions and decoded into sign pose sequences using the VAE decoder.}
  \label{fig:disentangled_vae_arch}
\end{figure*}

We adopt a two-stage framework in which continuous sign pose sequences are first encoded into a structured latent space using a disentangled Variational Autoencoder (VAE). 
Following the Gaussian latent formulation introduced in non-autoregressive SLP \cite{hwang2021non}, we explicitly disentangle the latent space across articulators by assigning dedicated subspaces to the right hand, left hand, body, and face. Such articulator-wise disentanglement has previously been shown to be effective in sign language production, as it prevents dominant body motion from masking fine-grained manual and facial articulation \cite{tasyurek2025disentangle}. In contrast to prior deterministic formulations, our encoder maps each pose frame to articulator-wise latent distributions rather than point embeddings, enabling distributional modeling without point-wise latent sampling at the encoding stage.

We represent sign poses as 3D skeleton sequences with $(x, y, z)$ coordinates, including the upper body (8 joints), left and right hands (21 joints each), and face (128 keypoints), consistent with \cite{tasyurek2025disentangle}. The VAE encoder is anatomically partitioned into four regions, each processed by a 2-layer residual linear encoder. The total latent dimensionality is fixed to 80, allocated as 8 for the upper body, 28 for each hand, and 16 for the face. Separate linear heads predict the latent mean $\mu$ and log-variance $\log\sigma^2$ for each articulator.

Each articulator is flattened per frame and encoded independently using a region-specific multi-layer perceptron (MLP). The input dimensionalities for the upper body, right hand, left hand, and face branches are $8 \times 3$, $21 \times 3$, $21 \times 3$, and $128 \times 3$, respectively. Each encoder branch outputs a latent representation that is linearly projected to region-specific mean and log-variance vectors of matching dimensionality.

Latent variables are sampled using the reparameterization trick,
\[
\mathbf{z}_t^{(a)} = \boldsymbol{\mu}_t^{(a)} + \boldsymbol{\sigma}_t^{(a)} \odot \boldsymbol{\varepsilon}, \quad \boldsymbol{\varepsilon} \sim \mathcal{N}(\mathbf{0}, \mathbf{I}),
\]
where $a \in \{\text{RH}, \text{LH}, \text{Body}, \text{Face}\}$ denotes the articulator. The complete latent representation for each frame is obtained by concatenating all regional latent vectors, resulting in an 80-dimensional latent code per frame.

Given a pose sequence $\mathbf{X}=\{x_t\}_{t=1}^{T}$, the encoder defines a factorized posterior distribution over articulator-specific latents:
\begin{equation}
q(z_t^{(a)} \mid x_t) = \mathcal{N}(\mu_t^{(a)}, \sigma_t^{2(a)}), \quad a \in \{\text{RH}, \text{LH}, \text{Body}, \text{Face}\}.
\end{equation}

In the base configuration, all regions are encoded using two-layer MLPs, with hidden dimensions set to twice the corresponding latent size: 16 for the upper body, 56 for each hand, and 32 for the face.

For CSL-Daily, increased encoder capacity is employed to better model its longer sequences and higher articulatory variability, consistent with observations in \cite{tasyurek2025disentangle}. Specifically, the upper body branch remains a two-layer MLP with hidden size $8 \times 3 = 24$, while the hand and face branches are extended to three-layer MLPs. The hand encoders use hidden dimensions of $28 \times 3 = 84$ and $28 \times 2 = 56$, and the face encoder uses hidden dimensions of $16 \times 3 = 48$ and $16 \times 2 = 32$.

Latent variables are sampled via the reparameterization trick and decoded using symmetric, structurally disentangled decoders to reconstruct pose coordinates. The VAE is trained using an articulator-weighted reconstruction loss to emphasize the refinement of hand articulations relative to other body parts together with a KL divergence term regularizing the latent distributions toward a standard Gaussian. A low-weight KL strategy is adopted to preserve reconstruction fidelity while maintaining structured latent representations. For CSL-Daily, the encoder and decoder capacity is increased to accommodate higher articulatory variability as in \cite{tasyurek2025disentangle}.

\subsection{Alignment-Aware Transformer with Distributional Supervision}
\label{sec:transformer_gloss}

After training the structurally disentangled VAE, its encoder is frozen and used to extract per-frame articulator-wise latent means and log-variances from ground-truth pose sequences. These latent statistics serve as supervision targets for the text-to-pose A$^{2}$V-SLP model. A non-autoregressive Transformer predicts these latent distribution parameters directly from text, instead of regressing pose coordinates. The encoder processes contextualized word representations obtained from BERT, and the decoder outputs for each target frame $t$ a concatenated vector of predicted mean and log-variance:
\begin{equation}
\hat{z}_t = [\hat{\mu}_t \;\Vert\; \hat{\log \sigma_t^2}],
\end{equation}
which is subsequently split into articulator-specific components.

Input text is represented using 768-dimensional contextualized word embeddings extracted from a pretrained BERT model. These embeddings are linearly projected to a 512-dimensional space to match the internal model dimensionality. Positional encoding is applied to preserve the temporal ordering of the text sequence. The text encoder is implemented as a Transformer encoder with 3 layers, 4 attention heads, and 1024-dimensional feed-forward networks. It produces contextualized token-level representations that serve as keys and values for cross-attention in the decoder. A lightweight length predictor operates on the mean-pooled encoder outputs to estimate the normalized target sequence length.

The decoder follows a non-autoregressive Transformer architecture to mitigate error accumulation during long sequence generation and to enable faster, fully parallel inference \cite{tasyurek2025disentangle}. It consists of 6 layers with 8 attention heads and 1024-dimensional feed-forward networks. Rather than relying on previously generated outputs, the decoder operates on a fixed set of learned time queries that define a temporal scaffold for the target sequence. 

Time queries are initialized from a stationary reference pose corresponding to a neutral signing state in which both hands rest downward. The reference pose is flattened, projected into the decoder embedding space, and replicated across all target time steps. This initialization provides a consistent and physically plausible starting point for pose generation while remaining independent of previously generated frames \cite{tasyurek2025disentangle}.

To improve temporal alignment in a gloss-free setting, we integrate a gloss attention mechanism into the Transformer decoder. Inspired by gloss attention proposed for gloss-free sign language translation \cite{yin2023gloss}, this mechanism introduces locality as an inductive bias to partially replace the alignment role of gloss annotations. Instead of standard global decoder self-attention over all frames, each frame query attends only to a fixed-size local temporal window of $N$ neighboring frames. This restricts the temporal receptive field of the decoder and biases learning toward short-range temporal dependencies, which are dominant in continuous sign motion.

Our use of gloss attention differs from its original formulation for sign language translation. In SLT, gloss attention operates over visual frame features to improve video–text alignment. In contrast, we apply gloss attention to decoder-side self-attention over latent frame representations in the pose generation process. Meanwhile, cross-attention to the text encoder remains global, allowing sentence-level semantic information to influence the entire sequence. This separation keeps semantic guidance global while enforcing locally structured motion generation.


The Transformer is trained using L1 loss to regress the VAE-encoded latent means and log-variances. This distribution-level supervision preserves a compact latent representation while encoding articulatory variability through variance estimates, without stochastic sampling during training. At inference time, latent codes are sampled from the predicted distributions and decoded into pose sequences using the pretrained VAE decoder.


\subsection{Transformer Training Strategy}
\label{sec:training}

Transformer training is performed in two phases. In the first phase, the Transformer is optimized to regress VAE-encoded latent parameters using masked L1 losses:
\begin{equation}
\mathcal{L}_{\text{latent}} = \sum_a \lambda_a \left(
\lVert \hat{\mu}^{(a)} - \mu^{(a)} \rVert_1 +
\lVert \hat{\log \sigma}^{2(a)} - \log \sigma^{2(a)} \rVert_1
\right),
\end{equation}
where $\lambda_a$ are articulator-specific weights. To better preserve fine-grained manual articulation, we introduce a dynamic L1 weighting strategy for hand joints. Base weights are assigned to right and left hands and dynamically increased up to a predefined maximum factor. This allows the model to emphasize expressive regions without destabilizing training.

In the second phase, KL regularization is enabled between the predicted latent distributions and the VAE posterior distributions:
\begin{equation}
\mathcal{L}_{\text{KL}} = \sum_a \text{KL}\big(
\mathcal{N}(\hat{\mu}^{(a)}, \hat{\sigma}^{2(a)})
\;\|\;
\mathcal{N}(\mu^{(a)}, \sigma^{2(a)})
\big).
\end{equation}

This stage aligns the distribution predicted by the Transformer with the VAE posterior. Following the structurally disentangled distributional training strategy introduced in \cite{tasyurek2025disentangle}, the final objective combines latent regression, KL regularization, and sequence length prediction loss.

\section{EXPERIMENTS}
\label{sec:experiments}

\subsection{Datasets}

We evaluate our approach on two widely used continuous sign language benchmarks with distinct linguistic and visual characteristics: PHOENIX-2014T and CSL-Daily. 

PHOENIX-2014T \cite{8578910} consists of 8,247 German Sign Language (DGS) sentences from weather broadcasts, aligned with gloss annotations and spoken German translations. We use the 3D pose annotations released in the CVPR 2025 SLRTP Challenge \cite{walsh2025slrtp}, where 2D keypoints extracted with MediaPipe Holistic \cite{lugaresi2019mediapipeframeworkbuildingperception} covering the upper body, both hands, and the face, and are subsequently lifted to 3D using Ivashechkin et al.’s method \cite{ivashechkin2023improving3dposeestimation}. Each sequence is represented as an $N \times 178 \times 3$ tensor, with $N$ frames and 178 concatenated landmarks.

CSL-Daily \cite{zhou2021signbt} contains 20,654 continuous Chinese Sign Language (CSL) sentences from 10 signers and covers a broader range of daily-life contexts, resulting in higher semantic and motion variability and making it a more challenging benchmark. Since CSL-Daily does not provide 3D pose annotations, we extract 3D skeletal representations directly using MediaPipe Holistic and standardize them to the same 178-keypoint format. To ensure consistency across datasets, poses are centered at the neck and scaled by shoulder width to improve invariance to signer and camera differences while preserving relative articulator motion.

\subsection{Evaluation Protocol}
Linguistic fidelity is evaluated via back-translation: generated pose sequences are translated back to spoken language using a pretrained sign-to-text model following \cite{walsh2024signstitchingnovelapproach}. We report BLEU, ROUGE, and chrF between back-translated outputs and reference sentences. Motion quality is assessed using DTW-MJE \cite{10.1145/3474085.3475463}, which measures spatio-temporal joint error.

\subsection{VAE Training}

In the first stage of the framework, the variational autoencoder is trained using a $\beta$-VAE objective,
\begin{equation}
\mathcal{L} = \mathcal{L}_{\text{rec}} + \beta\, \mathcal{D}_{\text{KL}},
\end{equation}
where $\mathcal{L}_{\text{rec}}$ denotes the reconstruction loss, computed as a weighted sum over articulator-specific terms,
\begin{equation}
\mathcal{L}_{\text{rec}} =
w_{\text{RH}} \mathcal{L}_{\text{RH}} +
w_{\text{LH}} \mathcal{L}_{\text{LH}} +
w_{\text{F}}  \mathcal{L}_{\text{F}} +
w_{\text{B}}  \mathcal{L}_{\text{B}},
\end{equation}
and $\mathcal{D}_{\text{KL}}$ enforces regularization of the latent distribution toward the prior.

The weighting coefficient $\beta$ controls the trade-off between reconstruction fidelity and latent regularization. Large values of $\beta$ over-constrain the latent space, limiting the model’s ability to reconstruct fine-grained pose details. To preserve articulatory precision, particularly in hand motion, we adopt a low KL weighting and set $\beta = 10^{-6}$ for PHOENIX-2014T and $\beta = 10^{-7}$ for CSL-Daily.

Region-specific reconstruction weights are used to reflect the varying perceptual importance of different articulators, with $w_{\text{RH}}=10$, $w_{\text{LH}}=14$, $w_{\text{F}}=2$, and $w_{\text{B}}=1$ for the right hand, left hand, face, and upper body, respectively.

The model is optimized using the Adam optimizer with a learning rate of $2 \times 10^{-4}$ and momentum parameters 0.5 and 0.9. Training is performed for 300 epochs.



\subsection{Generator Training and Evaluation}

After training the structurally disentangled VAE, we freeze its encoder and use it to extract per-frame articulator-wise latent statistics from ground-truth pose sequences. Specifically, for each frame $x_t$, the VAE encoder provides $(\mu_t, \log\sigma_t^2)$ (concatenated across articulators), which we treat as distributional supervision targets for our text-to-pose A$^{2}$V-SLP model.

We adopt a two-phase training scheme to stabilize learning and preserve structured latent distributions. In the first phase, the Transformer is trained to predict articulator-wise latent distribution parameters obtained from the pretrained disentangled VAE encoder. We minimize an L1 loss between predicted and VAE-encoded statistics, together with supervision on the target length ratio as in \cite{tasyurek2025disentangle}. In the second phase, we introduce a KL divergence term to regularize the predicted latent distributions, encouraging alignment with the empirical articulator-wise priors estimated from ground-truth pose encodings.

We employ the Adam optimizer with a learning rate of \(2 \times 10^{-4}\), weight decay of \(1 \times 10^{-4}\), and a ReduceLROnPlateau scheduler (factor 0.9, patience 40) in both phases of the training. Early stopping based on validation loss is applied to prevent overfitting. Training is performed on a 4×NVIDIA A100-SXM4-40GB setup using PyTorch Lightning.

\subsubsection{Structural Disentanglement and Gloss Attention}
\label{sec:gloss-attention-experiments}

We first analyze the effect of structurally disentangled latent representations and gloss attention. Tables~\ref{tab:dis-ablation} and~\ref{tab:gloss-ablation} show that articulator-wise latent disentanglement outperforms entangled representations, and that replacing global decoder self-attention with gloss attention yields substantial gains across all metrics, underscoring the importance of region-specific latent modeling and locality constraints for accurate temporal alignment in gloss-free SLP.

\begin{table}[h]
\centering
\scriptsize
\setlength{\tabcolsep}{4pt}
\begin{tabular}{lccccccc}
\toprule
\textbf{Model} & \textbf{B1↑} & \textbf{B2↑} & \textbf{B3↑} & \textbf{B4↑} & \textbf{chrF↑} & \textbf{ROUGE↑} \\
\midrule
GT (dev) & 30.92 & 21.34 & 16.01 & 12.74 & 34.71 & 30.19 \\
\midrule
Entangled (no split) & 23.23 & 14.31 & 9.85 & 7.38 & 28.70 & 21.18 \\
Disentangled (H-B-F) & \textbf{23.95} & \textbf{15.01} & \textbf{10.57} & \textbf{8.05} & \textbf{29.28} & \textbf{22.17} \\						
\bottomrule
\end{tabular}
\caption{Effect of disentanglement in the latent space.}
\label{tab:dis-ablation}
\end{table}	

\begin{table}[h]
\centering
\scriptsize
\setlength{\tabcolsep}{3.5pt}
\begin{tabular}{lccccccc}
\toprule
\textbf{Model} & \textbf{B1↑} & \textbf{B2↑} & \textbf{B3↑} & \textbf{B4↑} & \textbf{chrF↑} & \textbf{ROUGE↑} \\
\midrule
GT (dev) & 30.92 & 21.34 & 16.01 & 12.74 & 34.71 & 30.19 \\
\midrule
w/ Global Self-Attention & 23.95 & 15.01 & 10.57 & 8.05 & 29.28 & 22.17 \\
w/ Gloss Attention ($N{=}3$) & \textbf{30.09} & \textbf{20.12} & \textbf{14.76} & \textbf{11.59} & \textbf{33.42} & \textbf{29.15} \\						
\bottomrule
\end{tabular}
\caption{Effect of gloss attention mechanism used in place of transformer-based decoder's self-attention.}
\label{tab:gloss-ablation}
\end{table}	

\subsubsection{Gloss Attention Ablations}
\label{sec:gloss-attention-experiments}


We ablate the locality window size $N$, query construction strategy, and local--global attention fusion in gloss attention.

\begin{table}[h]
\centering
\scriptsize
\setlength{\tabcolsep}{5pt}
\begin{tabular}{lcccccc}
\toprule
\textbf{Locality} & \textbf{B1↑} & \textbf{B2↑} & \textbf{B3↑} & \textbf{B4↑} & \textbf{chrF↑} & \textbf{ROUGE↑} \\
\midrule
GT (dev) & 30.92 & 21.34 & 16.01 & 12.74 & 34.71 & 30.19 \\
\midrule
$N{=}3$ & \textbf{30.09} & \textbf{20.12} & \textbf{14.76} & \textbf{11.59} & \textbf{33.42} & \textbf{29.15} \\
$N{=}5$ & 28.52 & 18.85 & 13.60 & 10.54 & 32.54 & 27.68 \\
$N{=}7$ & 28.80 & 19.65 & 14.59 & 11.48 & 33.47 & 28.54 \\
$N{=}9$ & 28.01 & 18.29 & 13.11 & 10.12 & 32.21 & 26.81 \\
$N{=}11$ & 28.32 & 18.85 & 13.61 & 10.55 & 32.70 & 27.28 \\
\bottomrule
\end{tabular}
\caption{Effect of locality window size $N$ in gloss attention. Query type is "none".}
\label{tab:gloss_locality}
\end{table}


Table~\ref{tab:gloss_locality} shows that small locality windows perform best, with $N{=}3$ yielding the strongest results and larger windows degrading performance. Table~\ref{tab:gloss_query} evaluates query aggregation and local–global attention fusion under this best locality setting. Here, \textit{none} denotes frame-wise queries without temporal aggregation, \textit{mean} queries aggregate $q$ consecutive frames via average pooling, and \textit{attention} queries use a learnable self-attention module for aggregation. Attention-based aggregation degrades performance, while mean aggregation provide no consistent gain. In addition, experiments with larger locality windows ($N{>}3$) and different aggregation spans ($q$) did not yield further improvements. Frame-wise queries already offer sufficient temporal specificity, and aggregation tends to blur fine-grained temporal structure. Similarly, combining local gloss attention with global self-attention underperforms the local-only setting. These suggest that strong locality constraints are essential in decoder-side generation and that local gloss attention alone provides a more effective inductive bias for sign language production.
All gloss-attention ablations are conducted using fixed L1 reconstruction weights of $(14,10,2)$ for the right hand, left hand, and face, respectively, following prior work. We next investigate adaptive reconstruction loss weighting that adjust hand emphasis during optimization.

\begin{table}[h]
\centering
\scriptsize
\setlength{\tabcolsep}{2pt}
\begin{tabular}{lcccccc}
\toprule
\textbf{Variant} & \textbf{B1↑} & \textbf{B2↑} & \textbf{B3↑} & \textbf{B4↑} & \textbf{chrF↑} & \textbf{ROUGE↑} \\
\midrule
Local gloss (none, $N{=}3$) & \textbf{30.09} & \textbf{20.12} & \textbf{14.76} & \textbf{11.59} & \textbf{33.42} & \textbf{29.15} \\
\midrule
Attention query ($q{=}3$, $N{=}3$) & 28.63 & 18.92 & 13.77 & 10.81 & 32.53 & 27.70 \\
Mean query ($q{=}3$, $N{=}3$) & \textbf{29.41} & \textbf{19.74} & \textbf{14.49} & \textbf{11.38} & \textbf{33.34} & \textbf{28.33} \\
Mean query ($q{=}5$, $N{=}3$) & 28.45 & 18.98 & 13.81 & 10.82 & 33.06 & 27.71 \\
\midrule
Weighted local+global (none, $N{=}3$) & 27.12 & 17.74 & 12.92 & 10.16 & 31.48 & 25.61 \\
\bottomrule
\end{tabular}
\caption{
Effect of query construction and local–global attention fusion in gloss attention ($N{=}3$).
}
\label{tab:gloss_query}
\end{table}


\subsubsection{Training Strategy Ablations}

To reflect the dominant semantic role and fine-grained nature of manual articulations, we assign higher weights to left and right hand reconstruction losses. Rather than using fixed weights we employ a dynamic loss weighting scheme for hand regions. Contributions of $\mathcal{L}_{RH}$ and $\mathcal{L}_{LH}$ are weighted as
\begin{equation}
\lambda_{RH}(t) = b_{RH}\, s(t), \qquad
\lambda_{LH}(t) = b_{LH}\, s(t),
\end{equation}
where $(b_{RH}, b_{LH})$ define fixed right–left hand base weights and $s(t)$ is a shared boost factor. The boost is updated during training using exponential moving averages of the unweighted hand loss $\overline{\mathcal{L}}_{hand}$ and non-hand loss $\overline{\mathcal{L}}_{other}$ as
\begin{equation}
s(t+1) = \mathrm{clip}\!\left(
s(t)\left(\frac{\overline{\mathcal{L}}_{hand}}{\overline{\mathcal{L}}_{other}+\epsilon}\right)^{\alpha},
\,1,\, s_{\max}
\right),
\end{equation}
ensuring gradual adaptation and bounded stability.


\begin{table}[h]
\centering
\scriptsize
\setlength{\tabcolsep}{3pt}
\begin{tabular}{llccccc}
\toprule
\textbf{Base (RH,LH)} & \textbf{$b_{max}$} & \textbf{$(\lambda^{max}_{RH},\lambda^{max}_{LH})$} &
\textbf{B1↑} & \textbf{B4↑} & \textbf{chrF↑} & \textbf{ROUGE↑} \\
\midrule
GT (dev) & -- & -- & 30.92 & 21.34 & 34.71 & 30.19 \\
\midrule
(1.4, 1.0) & 10 & (14,10) & 29.91 & 11.90 & 33.66 & 28.67 \\
\midrule
(3.5, 2.5) & 4  & (14,10) & \textbf{29.96} & \textbf{12.23} & 33.78 & \textbf{29.12} \\
(3.5, 2.5) & 6  & (21,15) & 28.85 & 11.01 & 32.92 & 27.81 \\
(3.5, 2.5) & 8  & (28,20) & 29.53 & 11.68 & 33.59 & 28.60 \\
(3.5, 2.5) & 10 & (35,25) & 29.91 & 12.14 & 33.71 & 28.91 \\
\midrule
(4.0, 2.0) & 4 & (16,8)  & 29.68 & 12.09 & \textbf{33.98} & 28.96 \\
(3.0, 3.0) & 4 & (12,12) & 27.86 & 10.43 & 32.22 & 26.87 \\
(2.5, 3.5) & 4 & (10,14) & 27.91 & 10.13 & 32.69 & 27.09 \\
\bottomrule
\end{tabular}
\caption{Dynamic L1 hand-weighting ablations. We report base weights (RH,LH), maximum boost $b_{max}$, and the resulting maximum effective hand weights $(\lambda^{max}_{RH},\lambda^{max}_{LH}) = (base_{RH},base_{LH})\cdot b_{max}$. Face weights are fixed at 2. Query type is set to ”none” and window size is 3.}
\label{tab:l1_weight_ablation}
\end{table}		

We ablate the maximum boost $s_{\max}$ and base hand weights $(b_{RH},b_{LH})$ under controlled settings. Varying $(b_{RH},b_{LH})$ with fixed sum evaluates the right–left hand ratio, while a low-mass initialization $(1.4,1.0)$ with higher $s_{\max}=10$ tests the effect of total hand mass under the same effective ceiling $(14,10)$. Despite sharing the same maximum, the low-mass setting underperforms, indicating that stronger hand priors are beneficial early in training. Moreover, dynamic weighting with $(3.5,2.5)$ consistently outperforms fixed weights at the same scale, showing the advantage of adaptive scheduling. Overall, as shown in Table~\ref{tab:l1_weight_ablation}, the most stable performance is achieved with dynamically bounded hand weights $(14,10)$.

\begin{table}[h]
\centering
\scriptsize
\setlength{\tabcolsep}{5pt}
\begin{tabular}{lccccccc}
\toprule
\textbf{Model} & \textbf{B1↑} & \textbf{B2↑} & \textbf{B3↑} & \textbf{B4↑} & \textbf{chrF↑} & \textbf{ROUGE↑} \\
\midrule
GT (dev) & 30.92 & 21.34 & 16.01 & 12.74 & 34.71 & 30.19 \\
\midrule
A$^{2}$V-SLP & 29.96 & 20.59 & 15.43 & 12.23 & 33.78 & 29.12 \\
A$^{2}$V-SLP + KL & \textbf{31.23} & \textbf{21.80} & \textbf{16.54} & \textbf{13.28} & \textbf{35.10} & \textbf{30.71} \\							
\bottomrule
\end{tabular}
\caption{Effect of incorporating KL divergence.}
\label{tab:KL-results}
\end{table}

Table~\ref{tab:KL-results} shows that incorporating KL divergence further improves performance, supporting the use of distribution-level alignment between predicted and VAE-encoded latent representations.

\section{TEST RESULTS}
\label{sec:results}

Table~\ref{tab:results} shows that A$^{2}$V-SLP exceeds the performance of all prior gloss-free methods and gloss-based systems on PHOENIX14T, demonstrating that combining articulator-wise variational modeling with alignment-aware gloss attention effectively closes the gap between gloss-free and gloss-based SLP.

\begin{table}[h]
\centering
\scriptsize
\setlength{\tabcolsep}{2.5pt}
\begin{tabular}{l|c c c c c c c}
\toprule
\textbf{Model} & \textbf{DTW↓} & \textbf{B1↑} & \textbf{B2↑} & \textbf{B3↑} & \textbf{B4↑} & \textbf{chrF↑} & \textbf{ROUGE↑}\\
\midrule
GT \textsuperscript{\dag} (test) & 0.000 & 34.43 & 22.08 & 16.13 & 12.81 & 34.62 & 35.22\\
\midrule
\multicolumn{8}{c}{\textit{Gloss-based methods}} \\ 
\midrule
PT Base \cite{walsh2024signstitchingnovelapproach} & 0.197 & 6.27 & 3.33 & 2.14 & 1.59  & 9.50 & 14.52 \\ 
PT FP\&GN \cite{walsh2024signstitchingnovelapproach} & 0.191 & 11.45 & 7.08 & 5.08 & 4.04 & 19.09 & 14.52 \\
G2P-MP \cite{xie2023g2pddmgeneratingsignpose} & 0.146 & 15.43 & 10.69 & 8.26 & 6.98 & - & - \\
G2P-DDM \cite{xie2023g2pddmgeneratingsignpose} & 0.116 & 16.11 & 11.37 & 9.22 & 7.50 & - & - \\
GEN-OBT \cite{10.1145/3503161.3547830} & - & 24.92 & 15.72 & 11.20 & 8.68 & - & 25.21 \\
GCDM \cite{10.1145/3663572} & - & 22.03 & 14.21 & 10.16 & 7.91 & 23.20 & - \\
NAT-AT \cite{10.1145/3474085.3475463} & 0.177 & 14.26 & 9.93 & 7.11 & 5.53 & 21.87 & 18.72 \\
NAT-EA \cite{10.1145/3474085.3475463} & 0.146 & 15.12 & 10.45 & 7.99 & 6.66 & 22.98 & 19.43 \\
SLRTP25-Winner \textsuperscript{\dag} \cite{walsh2025slrtp} & \textbf{0.045} & \textbf{34.85} & \textbf{21.96} & \textbf{15.65} & \textbf{12.06} & \textbf{36.83} & \textbf{36.59} \\
\midrule
\multicolumn{8}{c}{\textit{Gloss-free methods}} \\
\midrule
SignVQNet \cite{hwang2024autoregressivesignlanguageproduction} & 0.299 & - & - & - & 6.88 & - & - \\
Stitching T2P \cite{walsh2024signstitchingnovelapproach} & 0.572 & 25.14 & 13.54 & 8.98 & 6.67 & 29.50 & 26.49 \\
NSLP-G \cite{hwang2024autoregressivesignlanguageproduction} & 0.638 & - & - & - & 5.56 & - & - \\
NSLP-G w/o FT \cite{hwang2024autoregressivesignlanguageproduction} & 0.646 & - & - & - & 4.41 & - & - \\
NSVQ + Non-AR \cite{walsh2024datadrivenrepresentationsignlanguage} & 0.105 & 27.74 & 16.36 & 11.75 & 9.20 & - & 27.93 \\
Sign-IDD \cite{tang2024signiddiconicitydisentangleddiffusion} & - & 25.40 & - & - & 8.93 & - & 27.60 \\
DARSLP-KL \cite{tasyurek2025disentangle} \textsuperscript{\dag} & 0.039 & 33.17 & 20.47 & 14.38 & 11.07 & 31.69 & 32.55 \\
ILRSLP-KL \cite{Kiziltepe_2025_ICCV} \textsuperscript{\dag} & 0.037 & 32.72 & 20.69 & 14.78 & 11.37 & 34.18 & 34.10 \\
\textbf{A$^{2}$V-SLP (Ours)} \textsuperscript{\dag} & 0.037 & 34.76 & 21.91 & 15.66 & 12.17 & 33.60 & 35.02 \\
\textbf{A$^{2}$V-SLP + KL (Ours)} \textsuperscript{\dag} & \textbf{0.037} & \textbf{36.25} & \textbf{23.23} & \textbf{16.88} & \textbf{13.31} & \textbf{35.41} & \textbf{36.64} \\
\bottomrule
\end{tabular}
\caption{PHOENIX14T test set comparison. \textsuperscript{\dag} Scores are obtained using the back-translation model of \cite{walsh2025slrtp}.}
\label{tab:results}
\end{table}

\begin{figure}[h]
\centering

\begin{minipage}[t]{0.04\textwidth}
\vspace{1.0cm}  
\rotatebox{90}{\small a) ILRSLP}\\[0.5cm]
\rotatebox{90}{\small b) DARSLP}\\[0.5cm]
\rotatebox{90}{\small c) A$^{2}$V-SLP}\\[0.5cm]
\rotatebox{90}{\small d) Ground Truth}
\end{minipage}%
\begin{minipage}[t]{0.43\textwidth}
\vspace{0pt}
\includegraphics[width=\textwidth]{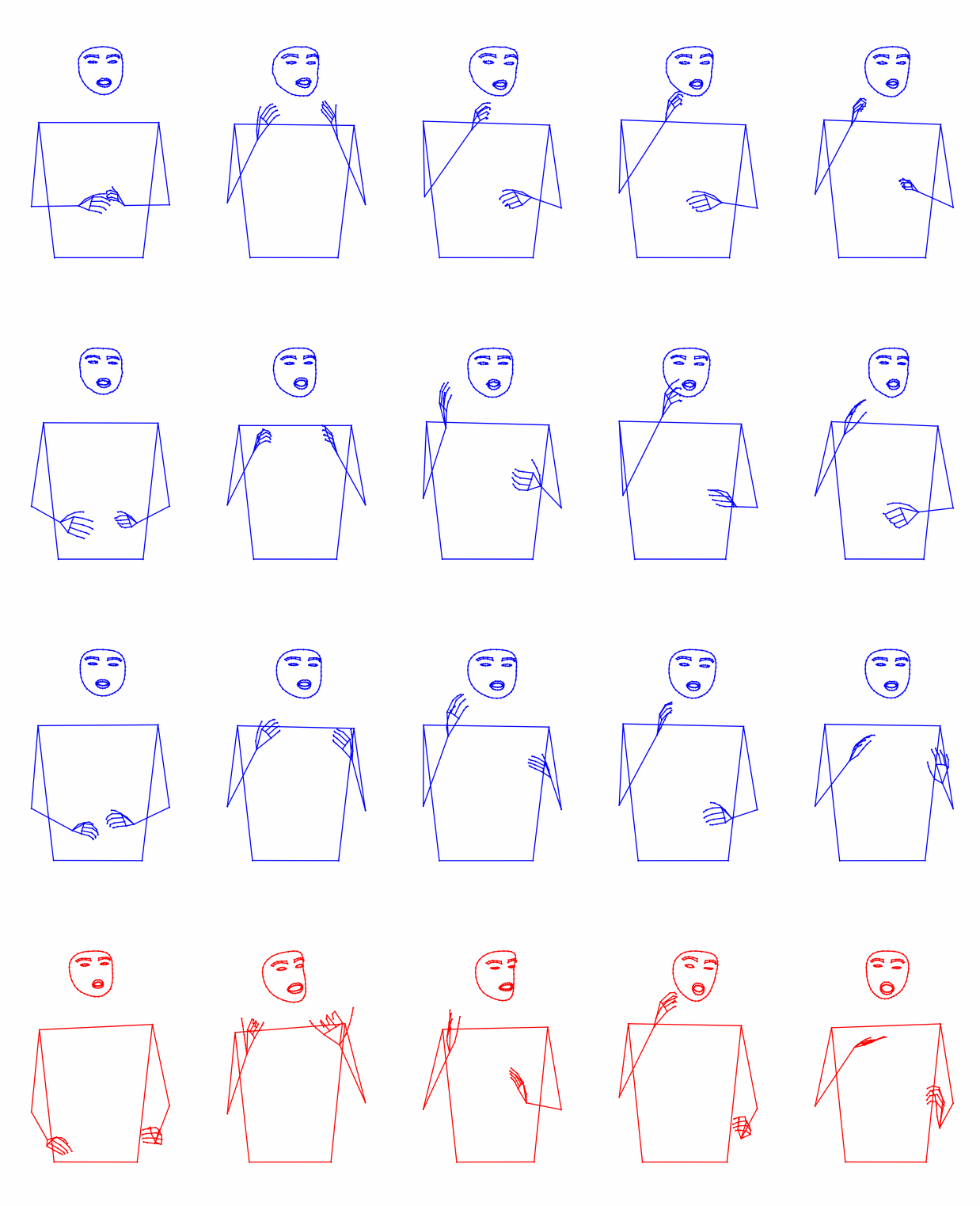}
\end{minipage}

\caption{\textbf{Pose sequence generated from input sentence:}
\textit{"am mittwoch zieht von der nordsee regen heran der sich am donnerstag ausbreitet"}}
\label{fig:pose_comparison}
\end{figure}

Qualitative examples in Figure~\ref{fig:pose_comparison} show that the proposed model captures key manual and non-manual articulations. Compared to IRSLP and DARSLP, A$^{2}$V-SLP produces pose sequences with improved temporal coherence and more faithful hand trajectories, particularly during transitions involving wrist rotation and finger articulation. The generated motions better preserve relative hand positioning. In addition, non-manual components such as head orientation and upper-body posture exhibit smoother temporal evolution and closer correspondence to the ground truth.

Using a non-autoregressive decoder, the model maintains smooth and plausible motion patterns without gloss supervision or large pretrained generative models, indicating effective performance under limited supervision.

\begin{table}[h]
    \centering
    \footnotesize
    \renewcommand{\arraystretch}{1.1}
    \setlength{\tabcolsep}{2pt} 
    \begin{tabular}{l|c c c c c c c}
        \toprule
        \textbf{Model} & \textbf{DTW↓} & \textbf{B1↑} & \textbf{B2↑} & \textbf{B3↑} & \textbf{B4↑} & \textbf{chrF↑} & \textbf{R↑}  \\
        \midrule
        GT (test) & 0.000 & 21.81 & 10.90 & 6.31 & 4.02 & 5.26 & 21.63 \\
        \midrule
        DARSLP \cite{tasyurek2025disentangle} & 0.177 & 18.72 & 8.33 & 4.46 & 2.73 & 4.24  & 19.13\\
        DARSLP+KL \cite{tasyurek2025disentangle} & 0.175 & 20.47 & \textbf{9.55} & \textbf{5.45} & \textbf{3.53} & \textbf{4.89}& \textbf{20.55} \\
        ILRSLP \cite{Kiziltepe_2025_ICCV} & 0.173 & \textbf{20.70} & 9.42 & 4.93 & 2.84 & 4.54 & 20.07\\
        ILRSLP+KL \cite{Kiziltepe_2025_ICCV} & 0.179 & 19.80 & 9.24 & 5.13 & 3.12 & 4.49 & 19.53\\
        A$^{2}$V-SLP (Ours) & \textbf{0.165} & 18.42 & 7.88 & 4.24 & 2.66 & 4.17 & 19.02 \\
        A$^{2}$V-SLP + KL (Ours) & 0.169 & 18.75 & 8.37 & 4.79 & 3.15 & 4.29  & 18.18 \\
        \bottomrule
    \end{tabular}
    \caption{Evaluation results on the CSL-Daily test set.}
    \label{tab:csl-results}
\end{table}

We evaluate the finalized model configuration, selected based on PHOENIX14T, on the CSL-Daily dataset. Due to the higher articulatory variability, the encoder–decoder capacity of the VAE is increased to three layers, following prior observations \cite{tasyurek2025disentangle}. The KL term is assigned a low weight ($10^{-6}$), as stronger regularization was found to dominate training; however, no CSL-specific hyperparameter optimization was performed. Table~\ref{tab:csl-results} shows that all models achieve closely clustered scores on CSL-Daily, reflecting the challenging nature of the dataset. Performance should be interpreted relative to the ground-truth (GT). While absolute gains are limited, A$^{2}$V-SLP attains the lowest DTW error, indicating improved motion alignment, and remains competitive with prior disentangled methods across linguistic metrics.

\section{CONCLUSIONS}
\label{sec:conlusion}

We introduced A$^{2}$V-SLP, an alignment-aware variational framework for gloss-free sign language production. The model predicts articulator-wise latent mean and variance parameters using a structurally disentangled VAE and a non-autoregressive Transformer, avoiding deterministic latent regression. Gloss attention is integrated into the decoder to guide temporal alignment between text and articulated motion without explicit gloss annotations. In addition, we employ adaptive, region-aware reconstruction weighting during training. Experimental results show consistent improvements in translation-based metrics and motion quality, highlighting the effectiveness of combining variational latent modeling, alignment-aware attention, and adaptive training strategies.

\section{Acknowledgements}
\label{Acknowledgements}
This research was supported by TÜBİTAK under the 1001 Program (Grant No. 124E618). We also acknowledge the EuroHPC Joint Undertaking for awarding us access to the Vega supercomputer at IZUM, Slovenia.
to emphasize fine-grained manual articulations 





{\small
\bibliographystyle{IEEEbib}
\bibliography{strings,refs}
}

\end{document}